\pdfoutput=1

\documentclass[11pt]{article}
\usepackage{amsmath,epsfig}

\DeclareMathOperator*{\argmax}{argmax}
\usepackage{multirow}
\usepackage{multicol}
\usepackage{amssymb}
\usepackage{subcaption}
\usepackage{bm}
\usepackage{cite}
\usepackage{booktabs}
\usepackage[colorlinks,linkcolor=blue]{hyperref}

\let\OLDthebibliography\thebibliography
\renewcommand\thebibliography[1]{
  \OLDthebibliography{#1}
  \setlength{\parskip}{0pt}
  \setlength{\itemsep}{0pt plus 0.3ex}
}

\usepackage{color}
\usepackage{algorithm}
\usepackage[noend]{algorithmic}
\usepackage{eqparbox}
\definecolor{applegreen}{rgb}{0.55, 0.71, 0.0}

\let\OLDthebibliography\thebibliography
\renewcommand\thebibliography[1]{
  \OLDthebibliography{#1}
  \setlength{\parskip}{0pt}
  \setlength{\itemsep}{0pt plus 0.3ex}
}

\usepackage{acl}
\usepackage{times}
\usepackage{latexsym}
\usepackage[T1]{fontenc}
\usepackage[utf8]{inputenc}
\usepackage{microtype}

\title{Overcoming Language Priors in Visual Question Answering via Distinguishing Superficially Similar Instances}

\author{Yike Wu\textsuperscript{1}\;\;\;Yu Zhao\textsuperscript{1}\;\;\;Shiwan Zhao \;\;\;Ying Zhang\textsuperscript{1}\thanks{\; Corresponding author.}  \\ {\bf Xiaojie Yuan\textsuperscript{1}\;\;\;Guoqing Zhao\textsuperscript{2}\;\;\;Ning Jiang\textsuperscript{2}} \\
\textsuperscript{1} Nankai University, Tianjin, China \;
\textsuperscript{2} Mashang Consumer Finance Co, Ltd \\
{\tt wuyike@nankai.edu.cn,} {\tt zhaoyu@dbis.nankai.edu.cn} \\
{\tt zhaosw@gmail.com}, {\tt \{yingzhang,yuanxj\}@nankai.edu.cn}
}

\begin{document}
\maketitle
\begin{abstract}
Despite the great progress of Visual Question Answering (VQA), current VQA models heavily rely on the superficial correlation between the question type and its corresponding frequent answers (i.e., language priors) to make predictions, without really understanding the input. In this work, we define the training instances with the same question type but different answers as \textit{superficially similar instances}, and attribute the language priors to the confusion of VQA model on such instances. To solve this problem, we propose a novel training framework that explicitly encourages the VQA model to distinguish between the superficially similar instances. Specifically, for
each training instance, we first construct a set that contains its superficially similar counterparts. 
Then we exploit the proposed distinguishing module to increase the distance between the instance and its counterparts in the answer space. 
In this way, the VQA model is forced to further focus on the other parts of the input beyond the question type, which helps to overcome the language priors. Experimental results show that our method achieves the state-of-the-art performance on VQA-CP v2. Codes are available at \href{https://github.com/wyk-nku/Distinguishing-VQA.git}{Distinguishing-VQA}.
\end{abstract}

\section{Introduction}

\begin{figure}[t]
     \centering
     \begin{subfigure}[b]{0.5\textwidth}
         \centering
         \includegraphics[width=\textwidth]{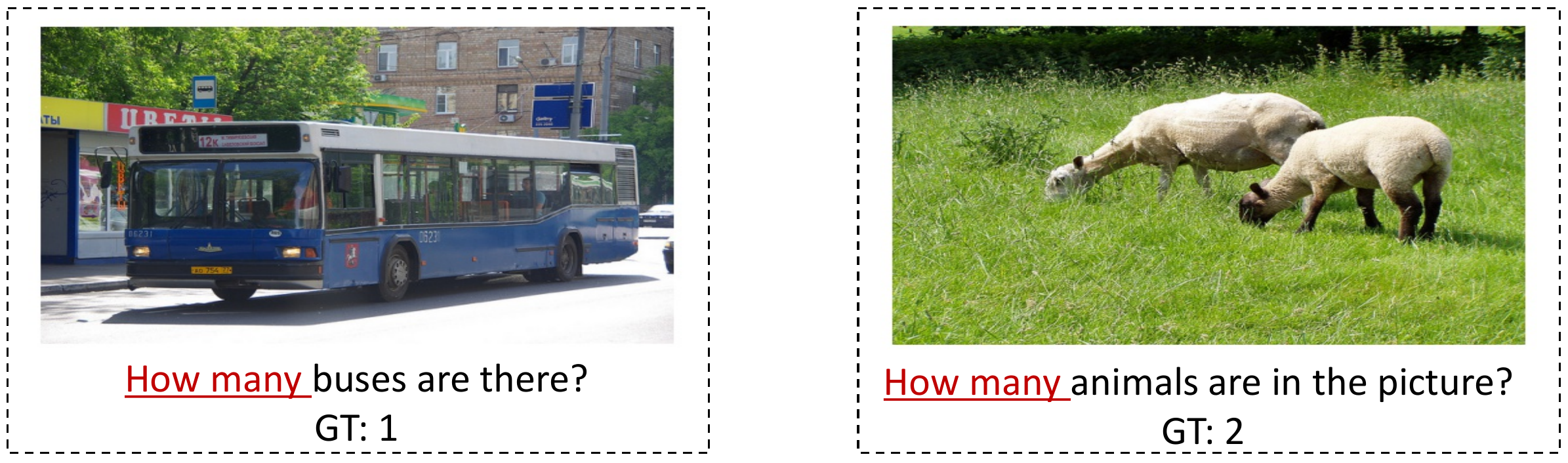}
         \caption{An example of superficially similar instances}
         \label{intro_a}
     \end{subfigure}
    \vspace*{\fill}
     \begin{subfigure}[b]{0.5\textwidth}
         \centering
         \includegraphics[width=\textwidth]{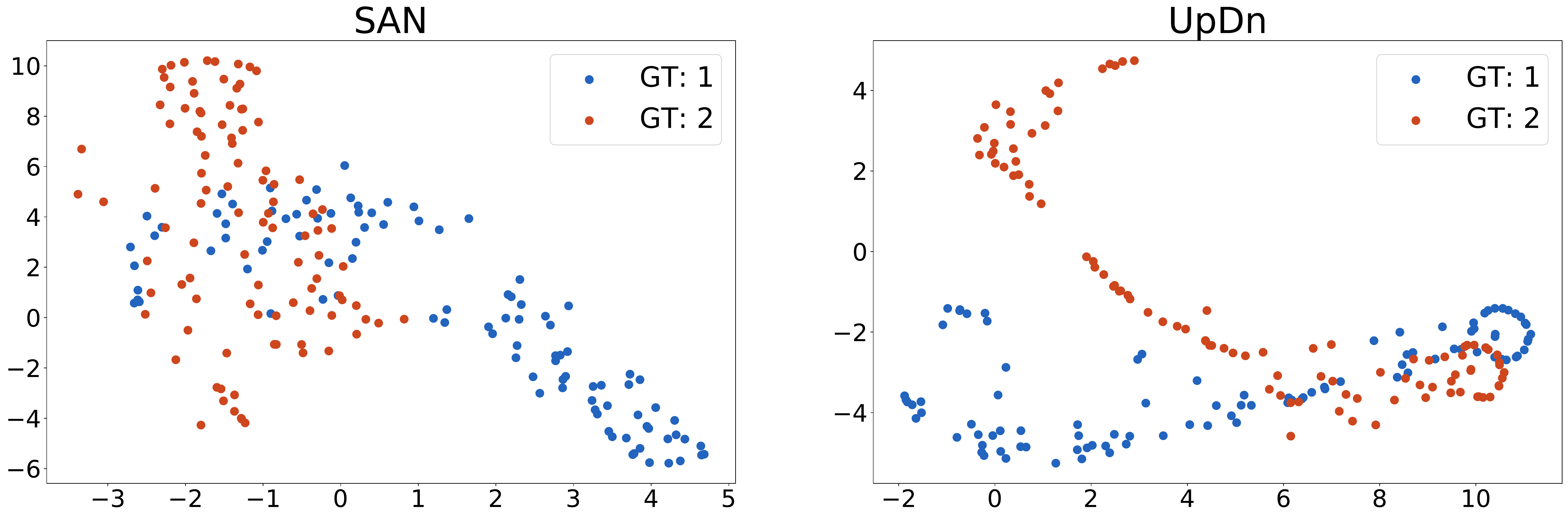}
         \caption{Superficially similar instances in the answer space}
         \label{intro_b}
     \end{subfigure}
     \caption{(a) displays an example of \textit{superficially similar instances}, which have the same question type (“How many”) but different ground-truth answers (“1” and “2”). (b) visualizes the projection of such instances in the answer space obtained by SAN \cite{san} and UpDn \cite{topdown} respectively. We observe that both models could not distinguish well between the mixed superficially similar instances.}
\end{figure}
Recent years have witnessed great progress in VQA based on deep learning. However, some researchers reveal that most existing VQA models heavily rely on the superficial correlation between the question type and its corresponding frequent answers to make predictions, instead of really understanding the input \cite{v2,analyzing,yinyang,assume}. For example, once a VQA model detects that a question begins with ‘‘how many’’, it tends to blindly output the most common answer ‘‘2’’ to the ‘‘how many’’ questions in the training data without looking at other parts of the input. 

We argue that the language priors arise from that the VQA model only captures the question type and ignores other information in the input image-question pair. On the one hand, the datasets which they are built upon are usually biased. For the questions of the same question type, the distribution of answers is severely biased to some frequent answers. On the other hand, deep learning methods tend to memorize some simple and salient patterns (e.g., frequency) in the training data, and easily exploit a shortcut to make predictions. Therefore, given an instance for testing, they prefer to directly look at the question type and leverage the superficial correlation rather than also analyze other information and further understand the whole input.  

For solving the above problem, we introduce the concept of \textit{superficially similar instances}, and propose to overcome the language priors via distinguishing between such instances. As shown in Figure \ref{intro_a}, the superficially similar instances refer to the instances that have the same question type but different answers.
By distinguishing such instances, the VQA model is forced to focus on other information besides the question type in the input image-question pair when making predictions. 

Unfortunately, it is not trivial to perform the distinguishing task, since a classical VQA paradigm does not have such a mechanism to support this. Given a training instance, the paradigm usually adopts the (binary) cross-entropy loss to reduce the distance between the projection of image-question pair in the answer space and the ground-truth, without explicitly encouraging the differentiation of superficially similar instances.
As shown in Figure \ref{intro_b}, the superficially similar instances are mixed up in the answer space, which indicates that both two widely-used VQA models \cite{san,topdown} could not distinguish between them well. 

In this work, we propose a novel training framework that explicitly encourages the VQA model to distinguish between the superficially similar instances. 
Specifically, for each training instance, we first construct a superficially similar set. The set consists of two kinds of superficially similar counterparts for the instance, which are collected in different ways and complement each other.
Then we exploit the designed \textit{distinguishing module} to increase the distance between a training instance and its superficially similar counterparts in the answer space.
Given a training instance, the distinguishing module urges the VQA model to give a higher probability on its ground-truth answer to itself than its superficially similar counterparts.
Finally, considering the cost of time and space, we implement our method in a resource-efficient way by manipulating the high-level features to construct the superficially similar set and sampling from the set.

In summary, the main contributions of this paper are as follows:
\begin{itemize}
    \item We are the first to introduce the concept of superficially similar instances and analyze the problem of language priors from this perspective. We also provide two different ways to collect superficially similar counterparts for a given instance.
    \item We propose a distinguishing module to explicitly encourage the differentiation between a training instance and its superficially similar counterparts, which forces the VQA model to further focus on other information in the input besides the question type. 
    \item Extensive experimental results demonstrate our approach successfully alleviates the language priors and really understands the input. Our method achieves the state-of-the-art results on the benchmark dataset VQA-CP v2, while maintaining competitive performance on the standard VQA v2 dataset. 
\end{itemize}

\section{Related Works} \label{related_works}

Despite the great progress in visual question answering\cite{san,topdown}, 
some researchers observe that most existing VQA models heavily rely on the language priors to make decisions \cite{analyzing,yinyang,v2}. Recently, \newcite{assume} propose a new split of the VQA v1 and VQA v2 datasets (VQA-CP v1 and VQA-CP v2 respectively), which makes the answer distribution of each question type different between the train and test splits. They find that the performance of existing VQA models drops significantly on their new splits compared to the original splits. This fully demonstrates the necessity of overcoming the language priors in VQA. 

Previous works on overcoming the language priors in VQA, which is called \textit{debiasing VQA methods}, can be roughly divided into four categories. 

\textbf{Methods modifying model architecture.} They usually design a specific model architecture to decompose the process of VQA into several steps. For instance, \newcite{assume} propose a Grounded Visual Question Answering (GVQA) model to disentangle the recognition of visual concepts from the identification of plausible answers. Similarly, \newcite{decompose} leverage the decomposed linguistic representations of different kinds of information in the question to decouple the discovery and verification of visual concepts. 

\textbf{Methods strengthening visual attention.} They usually leverage the human explanations (e.g., attention maps) to identify the important regions that are needed to answer the question correctly, and train a VQA model to focus on them. \newcite{hint} optimize the alignment between the human attention maps and the gradient-based importance of image regions from the VQA model. \newcite{scr} criticize the sensitivity of incorrect answers to the important regions that are identified based on human explanations. 

\textbf{Methods reducing unimodal bias.} They usually capture the unimodal biases from the language side with a question-only model, and propose strategies to reduce them. \newcite{adversary} train a VQA model and a question-only model that shares the same question encoder in an adversarial way.
\newcite{rubi} generate a 0-1 mask from the question-only model to modify the predictions of the VQA model, which modulates the importance of training instances with different levels of biases. Moreover, \newcite{lmh} train an ensemble of a VQA model and a pretrained question-only model, which prevents the VQA model from predicting answers in the way learned by the question-only model. \newcite{han2021greedy} capture different biases with multiple biased models in an ensemble manner, and reduce them step by step. Recently, some works also try to reduce the unimodal bias from a cause-effect perspective \cite{niu2021counterfactual,niu2021introd}.

\textbf{Methods balancing the dataset.} They usually make efforts to balance the dataset before training, which reduces the bias of the dataset itself.  \newcite{ssl} propose to balance the biased data without external annotations, and introduces an auxiliary task upon the balanced data to overcome the language priors. Additionally, \newcite{css} propose the CSS method to synthesize numerous counterfactual samples by masking critical objects in images or words in questions, and trains the VQA model on them to improve its visual-explainable and question-sensitive ability. Moreover, \newcite{liang2020learning} further improve the CSS method with a contrastive learning strategy.

Our approach shares the similar spirit with the methods balancing the dataset. However, previous works in this category only manipulate a single instance in different ways for data generation, while our method also considers the relationship between different instances in the dataset. And the proposed concept \textit{superficially similar instances} in this work is more general with wider coverage.

\begin{figure}[t]
\centering
\includegraphics[width=0.48\textwidth]{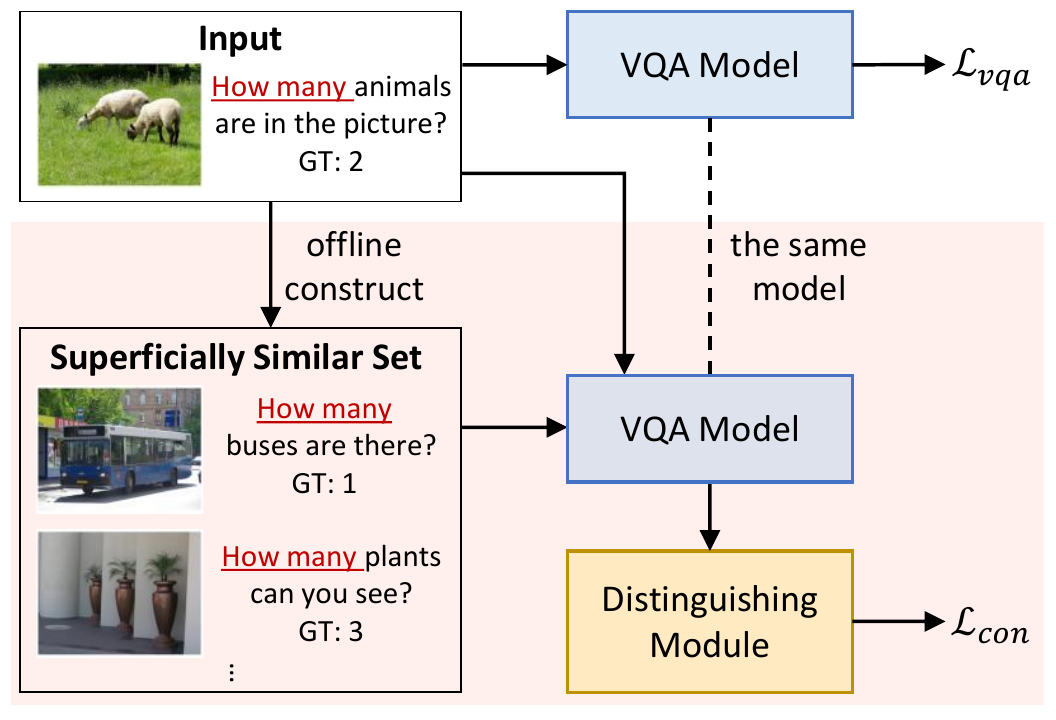}
\caption{The framework of our method. The pink part at the bottom shows the addition of our method compared to a classical VQA paradigm.}
\label{method}
\end{figure}

\section{Methodology}
The whole framework of our method is depicted in Figure \ref{method}. In the training phase, we optimize the whole framework with the loss $\mathcal{L}_{vqa}$ of a classical VQA paradigm and the proposed distinguishing loss $\mathcal{L}_{dis}$ in a multi-task manner, and the total loss $\mathcal{L}$ is defined as follows:
\begin{equation}
    \mathcal{L} = \lambda_1 \cdot \mathcal{L}_{vqa} + \lambda_2 \cdot \mathcal{L}_{dis}
\end{equation}
where $\lambda_1$ and $\lambda_2$ are coefficients tuning the influence of two losses. The distinguishing loss $\mathcal{L}_{dis}$ explicitly encourages the model to distinguish between the superficially similar instances. In the inference phase, we just utilize the trained VQA model to make decisions in a classical way.

Given a training instance, we compute its distinguishing loss $\mathcal{L}_{dis}$ in two steps. First, we offline construct a superficially similar set for the instance, which consists of its superficially similar counterparts. Second, when online training, we input the given instance with its superficially similar set into the VQA model, and further exploit the distinguishing module to compute $\mathcal{L}_{dis}$.

\subsection{Classical VQA Paradigm}
Given a VQA dataset $\mathcal{D} = \{(v_i, q_i, a_i)\}_{i=1}^{N}$ in which each instance is a triplet of an image $v_i \in \mathcal{V}$, a question $q_i \in \mathcal{Q}$, and the corresponding answer $a_i \in \mathbb{R}^{|\mathcal{A}|}$, a VQA paradigm typically learns a mapping from an image-question pair to its answer $\mathcal{F}: \mathcal{V} \times \mathcal{Q} \rightarrow \mathbb{R}^{|\mathcal{A}|}$, where $\mathcal{A}$ is the set of possible answers and each dimension of $a_i$ represents the confidence score of an answer in $\mathcal{A}$. The VQA paradigm can be instantiated with various models (e.g., the UpDn model \cite{topdown}).  
We follow the common formulation of the VQA task, which considers it as a multi-label classification problem. Given a training instance $(v_i, q_i, a_i) \in \mathcal{D}$, the VQA model takes the image-question pair $(v_i, q_i)$ as input, and output a vector of probabilities $p_i \in \mathbb{R}^{|\mathcal{A}|}$ over the answer set:
\begin{equation}
    p_{i} = \sigma(\mathcal{F}(v_i, q_i)),
\end{equation}
where $\sigma(\cdot)$ denotes the sigmoid function. The training objective is to minimize the binary cross-entropy loss between $p_i$ and $a_i$ for all instances in the dataset $\mathcal{D}$:
\begin{equation}
\begin{aligned}
    \mathcal{L}_{vqa} = & -\frac{1}{N}\sum_{i=1}^{N}\sum_{k=1}^{|\mathcal{A}|}(a_{ik} \cdot \mathrm{log} p_{ik} \\ & + (1 - a_{ik}) \cdot \mathrm{log}(1 - p_{ik}) ).
\end{aligned}
\end{equation}
The classical training objective only forces $p_i$ to approach $a_i$, but does not explicitly encourage the differentiation of superficially similar instances in the answer space. 

\subsection{Superficially Similar Set}
\label{set}
Before the VQA model learns to distinguish between superficially similar instances, we first construct a superficially similar set $\mathcal{S}$ for each training instance in the dataset $\mathcal{D}$. The set $\mathcal{S}$ is composed of two kinds of superficially similar counterparts for a given instance. 

\textbf{Real counterparts.} 
For the first kind, we directly select the existing instances in the training data with the same question type as the given instance but different answers, which we call \textit{real counterparts}. 

\textbf{Synthetic counterparts.} However, since the size of training data is limited, the real counterparts may be not sufficient for a VQA model to acquire satisfactory distinguishing ability. Therefore, we introduce the second kind of superficially similar counterparts as the complement, inspired by the previous work \cite{ssl}. We randomly select images from the training data, and combine them with the question of the given instance to construct new instances, which we call \textit{synthetic counterparts}. Obviously, these new instances and the given instance have the same question type (their questions are the same) but different answers\footnote{We ignore the rare case that the randomly selected image coincides with the information needed to get the same answer.} (their images are different). In this way, we can obtain much more superficially similar counterparts than the real counterparts. 

\textbf{Set construction.} The construction process of $\mathcal{S}$ is elaborated in Algorithm \ref{alg}, which is automatic and requires no additional annotation. Given a training instance $(v_i, q_i, a_i) \in \mathcal{D}$, we sequentially process all other training instances $\{(v_j, q_j, a_j)_{j \neq i} \in \mathcal{D} \}$ in two steps. First, we combine $q_i$ with $v_j$ to generate a synthetic counterpart superficially similar counterpart $(v_j, q_i)$, which have the same question type with $q_i$ but cannot be answered by $a_i$. Second, if the questions $q_i$, $q_j$ have the same question type and the corresponding answers $a_i$, $a_j$ are different, we also obtain a superficially similar real counterparts $(v_j, q_j)$. 

\begin{algorithm}[t]
\small
	\caption{Superficially Similar Set Construction}
	\label{alg}
	\begin{algorithmic}[1]
    \REQUIRE The VQA dataset $\mathcal{D}$; An instance $(v_i, q_i, a_i) \in \mathcal{D}$
	\ENSURE The superficially similar set $\mathcal{S}$ of $(v_i, q_i, a_i)$
	\STATE initialize an empty set $\mathcal{S}$
	\FOR{each training instance $(v_j, q_j, a_j)_{j \neq i} \in \mathcal{D}$} 
	\STATE  add ($v_j$, $q_i$) into $\mathcal{S}$ \COMMENT{Synthetic counterpart}
	\IF[The same question type]{type($q_{i}$) = type($q_{j}$)}
	\STATE $m \gets \argmax_{l \in \mathcal{A}} a_{il}$
	\STATE $n \gets \argmax_{l \in \mathcal{A}} a_{jl}$
    \IF[Different answers]{$m \neq n$}
	\STATE add ($v_j$, $q_j$) into $\mathcal{S}$ \COMMENT{Real counterpart}
	\ENDIF
	\ENDIF
	\ENDFOR
	\RETURN $\mathcal{S}$
    \end{algorithmic}
\end{algorithm}

\subsection{Distinguishing Module}
\label{cl module}
Once the superficially similar set of the training instance is constructed, we exploit the distinguishing module to explicitly encourage the distinction between the training instance and its superficially similar counterparts. Specifically, given a training instance $(v_i, q_i, a_i) \in \mathcal{D}$ with its superficially similar set $\mathcal{S}$, we first input them into the VQA model to get their projection in the answer space, i.e., a vector of probabilities $p_i \in \mathbb{R}^{|\mathcal{A}|}$ and a set of such vectors $\{p_j\}$ respectively. Next, the distinguishing module tries to increase the distance between $p_i$ and each element $p_j$ in $\{p_j\}$.

A naive solution is to directly increase the Euclidean distance between the two vectors $p_i$ and $p_j$. However, this is unreasonable in two folds. First, the direction along which the distance increases should be constrained. As $p_i$ keeps moving away from $p_j$ in the answer space, the value of its dimension corresponding with the ground-truth answer should increase simultaneously, which guarantees that the question $q_i$ can be answered correctly when the distance increases. Second, it doesn't make sense to increase the distance between $p_i$ and $p_j$ along other dimensions except the ones corresponding with their ground-truth answers. For example, if the answers to a training instance and its superficially similar counterpart is “2” and “3” respectively, it is meaningless to compare these two instances on other dimensions such as “yes”, “red”. 

\textbf{Distinguishing loss.} Considering the above two constraints, we design the distinguishing loss in Equation \ref{contrastive}. For each training instance $(v_i, q_i, a_i)$, the VQA model is encouraged to give higher probability on its ground-truth answer to itself than its superficially similar counterpart, and vice versa:

\begin{equation}
\begin{aligned}
    \mathcal{L}_{dis} = & - \frac{1}{N}\sum_{i=1}^{N}\sum_{j}(  \mathrm{log}\sigma(p_{im} - p_{jm})\\&+\mathrm{log}\sigma(p_{jn} - p_{in})),
    \label{contrastive}
\end{aligned}
\end{equation}
where $m$, $n$ are the ground-truth answer of $(v_i, q_i, a_i)$ and that of its superficially similar counterpart respectively. This training objective satisfies the mentioned two constraints. First, the direction of increasing the distance complies with the 
goal of answering the question correctly, e.g., the value of $p_{im}$ gets larger progressively. Second, the distance between $p_i$ and $p_j$ increases only along the dimensions corresponding with their ground-truth answers, e.g., raising the value of $(p_{im} - p_{jm})$ only increases the distance between $p_i$ and $p_j$ along the $m$-th dimension.

Note that the superficially similar relationship is symmetrical. The training instance $(v_i, q_i, a_i)$ is also the superficially similar counterpart of each instance in $\mathcal{S}$. Therefore, for the real counterparts from the dataset $\mathcal{D}$, the second term $\mathrm{log}\sigma(p_{jn} - p_{in})$ in Equation \ref{contrastive} can be omitted, since it has been included in $\mathcal{L}_{dis}$. On the other hand, the synthetic counterparts usually have no meaningful answer to define the second term. Thus we can simplify Equation \ref{contrastive} to:
\begin{equation}
    \mathcal{L}_{dis} = - \frac{1}{N}\sum_{i=1}^{N}\sum_{j}\mathrm{log}\sigma(p_{im} - p_{jm}).
    \label{nofocal}
\end{equation}
\textbf{Modulating factor.} However, we observe that the improvement of performance is limited if we train the VQA model as Equation \ref{nofocal}. We deduce that this is mainly because we treat each superficially similar counterpart equally and do not distinguish between hard and easy ones. More specifically, the superficially similar counterparts which yield a larger $p_{jm}$ are more difficult to distinguish from the training instance $(v_i, q_i, a_i)$, and should be penalized more in $\mathcal{L}_{dis}$. Therefore, inspired by the work \cite{focal}, we finally define the distinguishing loss as follows:
\begin{equation}
    \mathcal{L}_{dis} = - \frac{1}{N} \sum_{i=1}^{N}\sum_{j}p_{jm} \cdot \mathrm{log}\sigma(p_{im} - p_{jm}),
    \label{focal}
\end{equation}
where we add $p_{jm}$ as a \textit{modulating factor}. Consequently, $\mathcal{L}_{dis}$ will focus more on the hard superficially similar counterparts with larger $p_{jm}$ than the easy ones.

\begin{table*}[htbp]
\centering
\resizebox{1.0\textwidth}{!}{ 
\setlength{\tabcolsep}{1mm}{\begin{tabular}{lccccccccc}

\toprule
\multirow{ 2}{*}{Model} & \multicolumn{4}{c}{VQA-CP v2 test set} & \makebox[0.01\textwidth][c]{} & \multicolumn{4}{c}{VQA v2 val set}\\
\cline{2-5}
\cline{7-10}
& Overall & Yes/No & Num & Other & & Overall & Yes/No & Num & Other\\

\midrule
\multicolumn{10}{l}{\textit{classical methods}}\\
\midrule
SAN & 24.96 & 38.35 & 11.14 & 21.74 & & 52.02 & 68.89 & 34.55 & 43.80 \\
UpDn & 39.74 & 42.27 & 11.93 & 46.05 & & 63.48 & 81.18 & 42.14 & 55.66 \\

\midrule
\multicolumn{10}{l}{\textit{single-model debiasing methods}}\\
\midrule
GVQA & 31.30 & 57.99 & 13.68 & 22.14 & & 48.24 & 72.03 & 31.17 & 34.65 \\
AdvReg  & 41.17 & 65.49 & 15.48 & 35.48 & & 62.75 & 79.84 & 42.35 & 55.16\\
RUBi  & 47.11 & 68.65 & 20.28 & 43.18 & & 61.16 & - & - & - \\
HINT & 46.73 & 67.27 & 10.61 & 45.88 & & 63.38 & 81.18 & 42.99 & 55.56 \\
SCR & 49.45 & 72.36 & 10.93 & 48.02 & & 62.20 & 78.80 & 41.60 & 54.50 \\
DLR  & 48.87 & 70.99 & 18.72 & 45.57 & & 57.96 & 76.82 & 39.33 & 48.54\\
SSL  & \underline{57.59} & 86.53 & \underline{29.87} & \underline{50.03} & & 63.73 & - & - & - \\
Re-scaling & 47.09 & 68.42 & 21.71 & 42.88 & & 55.50 & 64.22 & 39.61 & 53.09 \\
CF-VQA   & 53.55 & \textbf{91.15} & 13.03 & 44.97 & & 63.54 & 82.51 & 43.96 & 54.30\\
\textbf{UpDn+DM (Ours)}  & \textbf{61.13\textsuperscript{$\pm$0.17}} & \underline{88.13\textsuperscript{$\pm$0.31}} & \textbf{45.98\textsuperscript{$\pm$0.63}} & \textbf{51.13\textsuperscript{$\pm$0.04}} & & 63.53\textsuperscript{$\pm$0.09} & 81.09\textsuperscript{$\pm$0.21} & 39.61\textsuperscript{$\pm$0.24} & 56.52\textsuperscript{$\pm$0.05}\\

\midrule
\midrule
\multicolumn{10}{l}{\textit{ensemble debiasing methods}}\\
\midrule

LMH   & 52.45 & 69.81 & 44.46 & 45.54 & & 61.64 & 77.85 & 40.03 & 55.04 \\
CSS  & 58.95 & 84.37 & 49.42 & 48.21 & & 59.91 & 73.25 & 39.77 & 55.11\\
CSS+CL  & 59.18 & 86.99 & 49.89 & 47.16 & & 57.29 & 67.27 & 38.40 & 54.71\\
CSS+LMH+Re-scaling & 56.55 & 83.95 & 47.81 & 44.59 & & 55.96 & 70.52 & 33.56 & 50.83 \\
GGE-DQ-tog  & 57.32 & 87.04 & 27.75 & 49.59 & & 59.11 & 73.27 & 39.99 & 54.39\\
CSS+IntroD  & 60.17 & 89.17 & 46.91 & 48.62 & & 62.57 & 78.57 & 41.42 & 56.00\\
\bottomrule
\end{tabular}}
}
\caption{Performance (\%) on VQA-CP v2 test set and VQA v2 val set.
\textbf{For fairness}, we mainly compare our approach with single-model methods, and highlight the obtained \textbf{best} and \underline{second best} results in each column.
We report the average with the standard variation of results with 5 random seeds.
}
\label{main results}
\end{table*}

\subsection{Resource-Efficient Implementation}
For the implementation of the proposed framework, we need to consider the efficiency of both time and space. On the one hand, given a batch of training data, it is time-consuming to run two respective passes in the VQA model to compute $\mathcal{L}_{vqa}$ and $\mathcal{L}_{dis}$. In practice, we only run one pass to extract the high-level features from a batch of image-question pairs, and share them during the computation of $\mathcal{L}_{vqa}$ and $\mathcal{L}_{dis}$. On the other hand, for a training instance, the GPU memory is always too limited to accommodate all the instances in its superficially similar set $\mathcal{S}$. In practice, we randomly sample $N_1$ the real counterparts and $N_2$ the synthetic counterparts from the intersection of $\mathcal{S}$ and the current batch of data when online training. Note that the sampling process is also at the feature level and the time cost is negligible.
The selection of $N_1$ and $N_2$ is discussed in Section \ref{discuss_n1n2}. 

% \vspace{-5pt}
\section{Experiments}
\subsection{Experimental Setup}
\textbf{Datasets and evaluation.} We evaluate the performance of the proposed method on the benchmark dataset VQA-CP v2 \cite{assume}. For completeness, we also evaluate our method on the standard VQA v2\cite{v2}. We use the standard VQA accuracy as the evaluation metric. 

\textbf{Implementation details.} We employ the UpDn model \cite{topdown} as the base architecture of our approach. The coefficient $\lambda_1$ and $\lambda_2$ are set to 0.05 and 0.6 respectively. The values of $N_1$ and $N_2$ are both set to 1. Additionally, we categorize the questions into 65 question types as in VQA v2 \cite{v2}.

\subsection{Quantitative Analysis}
\textbf{Baselines.} 
We compare the proposed method UpDn+DM with two classical methods SAN \cite{san} and UpDn \cite{topdown}, and other debiasing methods\footnote{Methods designed to overcome language priors.}:
(1) single-model methods: GVQA \cite{assume}, AdvReg \cite{adversary}, RUBi \cite{rubi}, HINT \cite{hint}, SCR \cite{scr}, DLR \cite{decompose}, SSL \cite{ssl}, Re-scaling \cite{guo2021loss}, CF-VQA \cite{niu2021counterfactual};
(2) ensemble methods: LMH \cite{lmh}, CSS \cite{css}, CSS+CL \cite{liang2020learning}, CSS+LMH+Re-scaling \cite{guo2021loss}, GGE-DQ-tog \cite{han2021greedy}, CSS+IntroD \cite{niu2021introd}.
However, the latter methods are based on stronger ensemble architecture. For fairness, we mainly compare with single-model methods.
\begin{figure*}[t]
    \centering
    \includegraphics[width=0.98\textwidth]{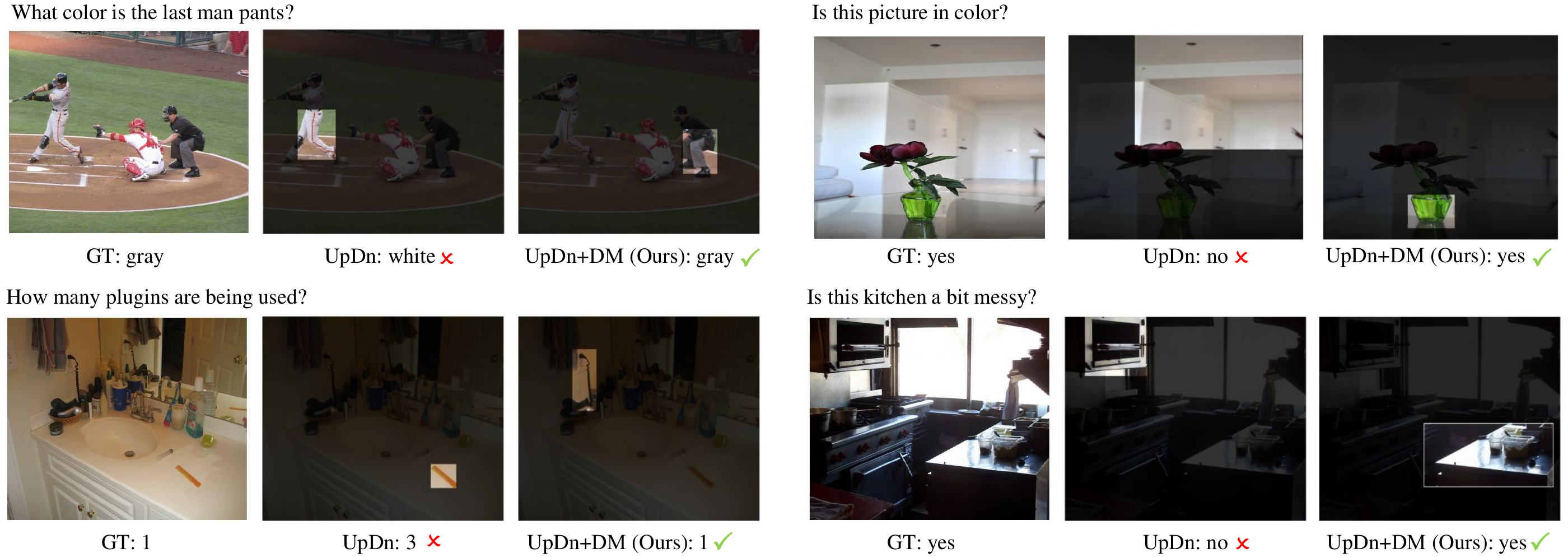}
    \caption{Qualitative examples in VQA-CP v2 test set. We display the ground-truth answer and predicted answers from UpDn and our method respectively. The best scoring region in the attention map when the model makes the prediction is highlighted in the image. 
    }
    \label{examples}
\end{figure*}

% In each example,
\begin{figure*}[t] 
  \centering 
  \subfloat[question type: “how many”]{
    \includegraphics[width=.31\textwidth]{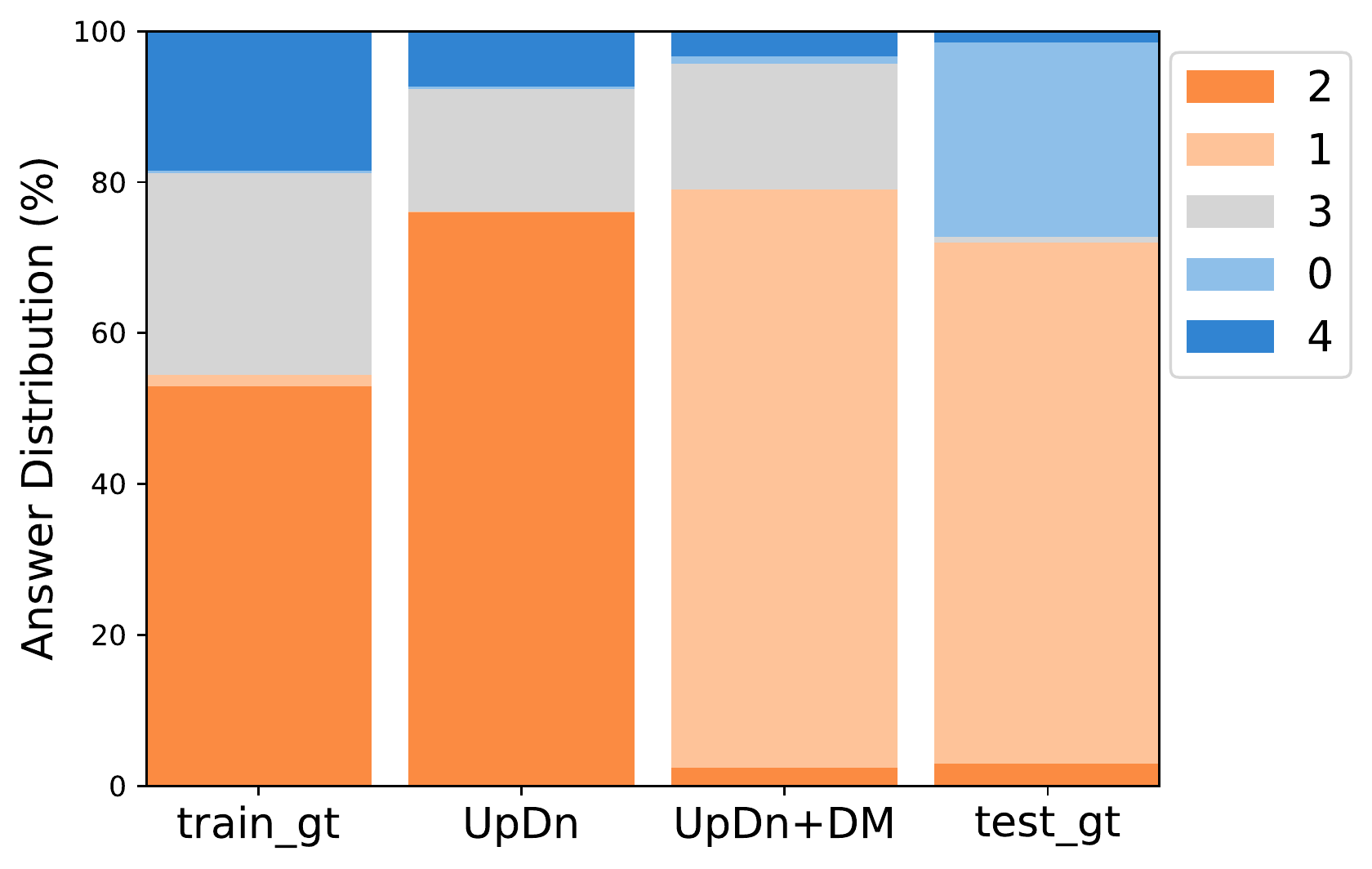}
    \label{reg_a}}
  \subfloat[question type: “what color is the”]{ 
    \includegraphics[width=.33\textwidth]{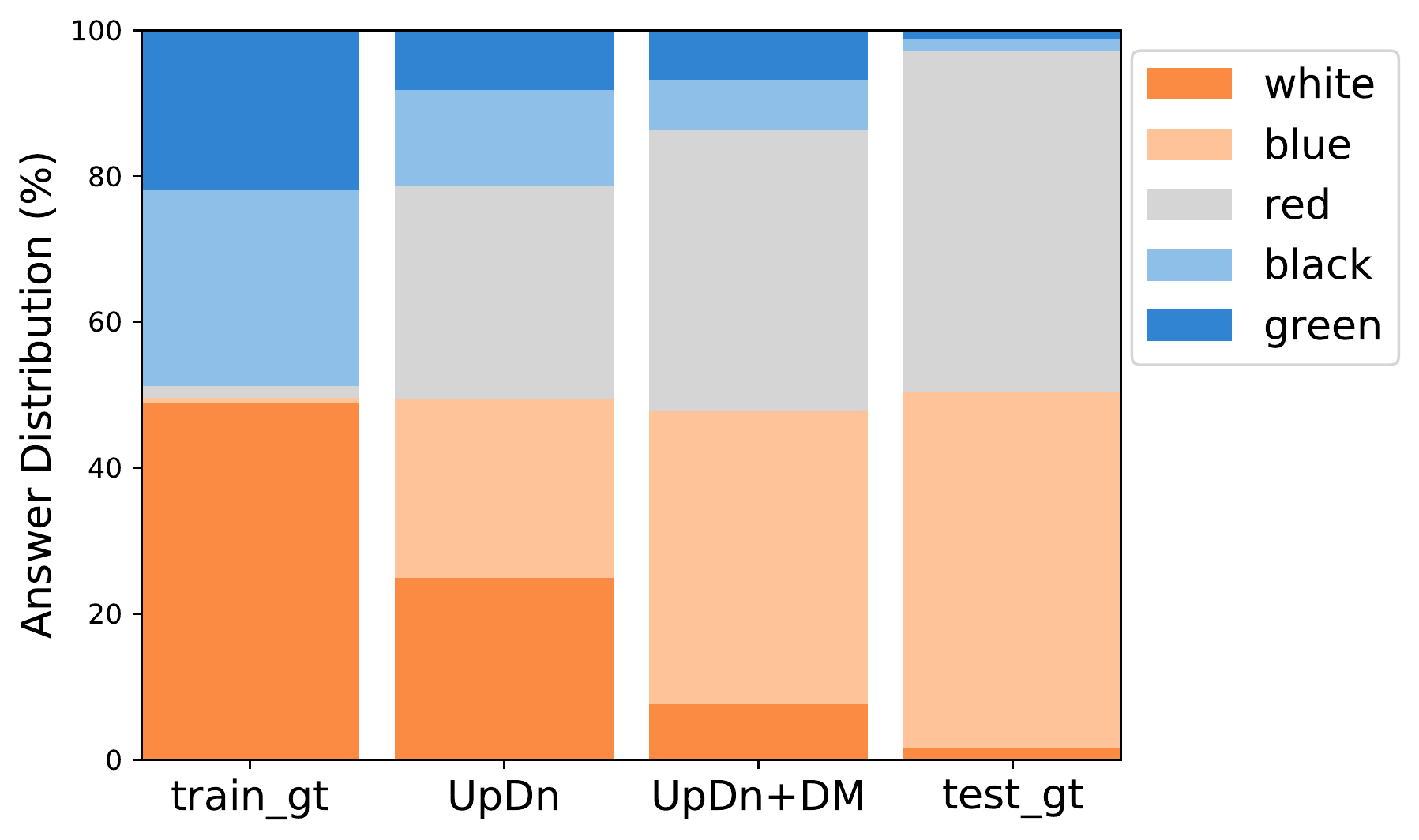}
    \label{reg_b}}
    \subfloat[question type: “is this”]{ 
    \includegraphics[width=.34\textwidth]{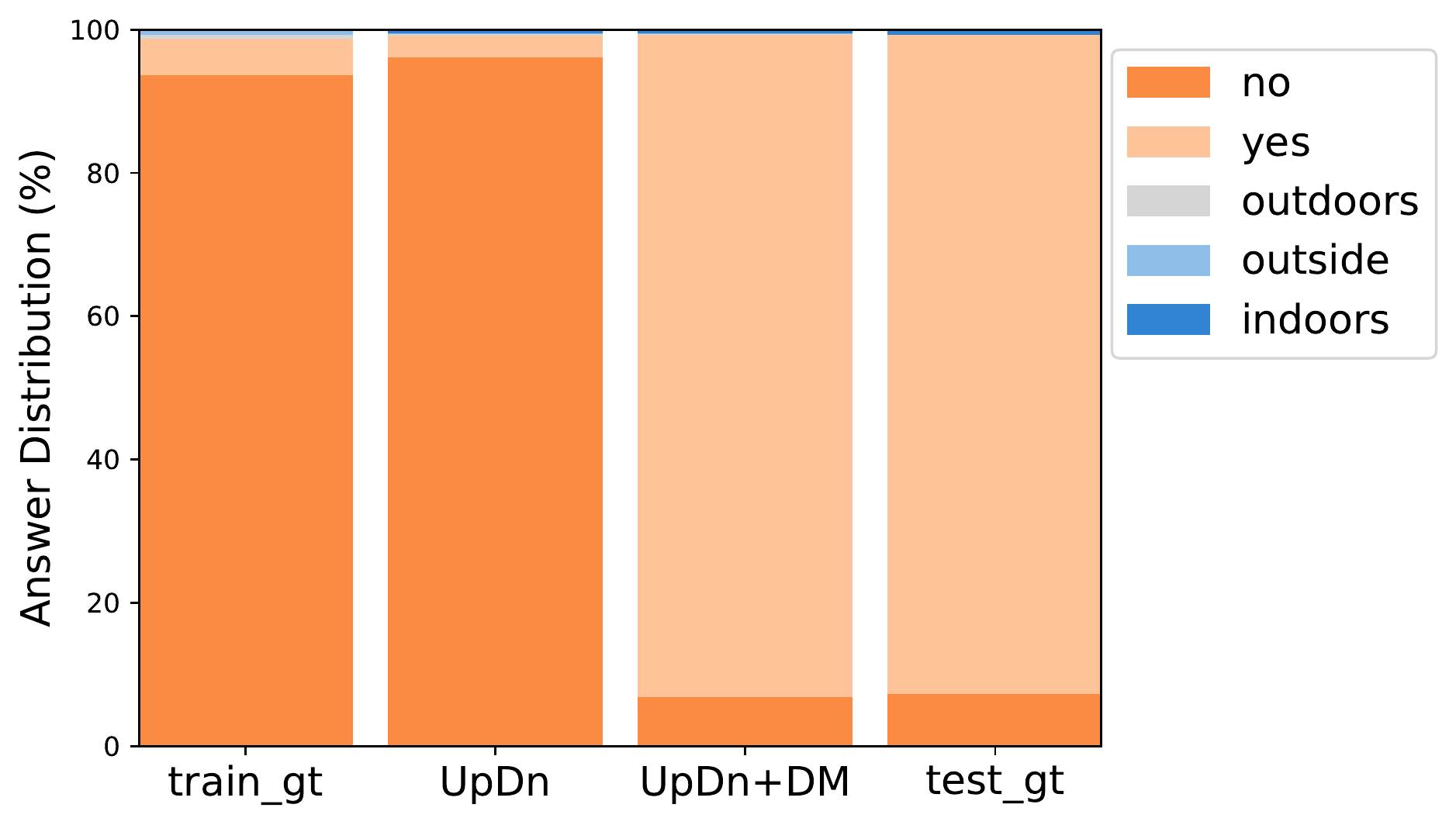}
    \label{reg_c}}
  \caption{The answer distributions in percentage (\%) of specific question types on VQA-CP v2 dataset. For each question type, we display the answer distributions for the ground-truth answers from the training set (“train-gt”), the baseline UpDn evaluated on the test set (“UpDn”), our method UpDn+DM evaluated on the test set (“UpDn+DM”) and the ground-truth answers from the test set (“test-gt”). Note that for readability, we only display the most frequent answers of the question type in VQA-CP v2 dataset and omit the others.}
  \label{answerdistribution} 
\end{figure*}
\begin{figure}[t]
    \centering
    \setlength{\belowcaptionskip}{-0.4cm}
    \includegraphics[width=0.5\textwidth]{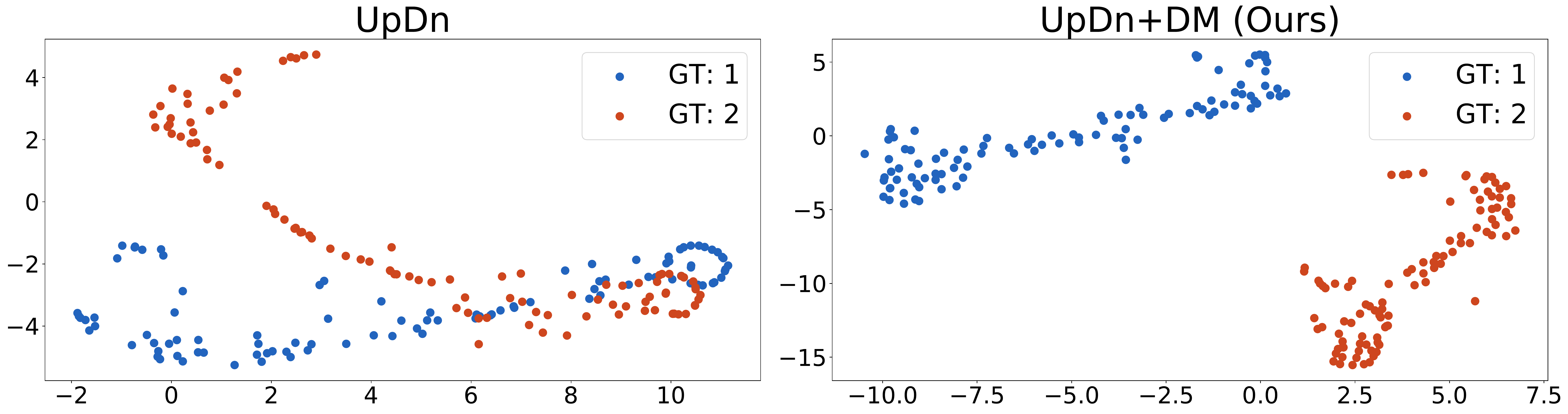}
    \caption{Visualization of the distance between superficially similar instances in the answer space using t-SNE. Each point denotes an instance with the same question type “how many”. Different colors denote different ground-truth answers.}
    \label{tsne}
\end{figure}

\textbf{Performance on VQA-CP v2 test set.} 
The results on VQA-CP v2 are reported in Table \ref{main results}. 
First, we observe that our approach outperforms classical methods by a large margin.
Second, our method performs best among single-model methods on the accuracies of “Overall”, “Yes/No” and “Other” categories, and second-best on the accuracy of "Yes/No". 
Moreover, even compared with stronger ensemble methods, our method also achieves competitive results. 
For example, our method outperforms the \underline{second-best} single-model method SSL by 3.54\% and the \textbf{best} ensemble-model method CSS-IntroD by 0.96\% on the overall accuracy.
It demonstrates the effectiveness of our approach at overcoming language priors. 

\textbf{Performance on VQA v2 val set.} 
The results on VQA v2 val set are reported in Table \ref{main results}. The debiasing VQA methods usually lead to a decline in the performance over VQA v2 val set, since they tend to penalize the superficial correlation excessively and over-correct the language priors. It is worth noting that our method maintains the competitive performance on VQA v2 with significant improvement on VQA-CP v2, which indicates the robustness of our method.

\subsection{Qualitative Analysis}
\textbf{Qualitative examples.} 
We display qualitative examples in Figure \ref{examples} to compare our method with its base model UpDn. 
We observe that UpDn+DM predicts the correct answers while UpDn does not, which shows the effectiveness of our method. 
For example,
as shown in Figure \ref{reg_b},
"white" is the most frequent answer in the training set with respect to the question type "what color is the".
While UpDn utilizes the language priors to locate the first man with a "white" shirt by mistake, 
our method is able to capture the right region "the last man's pants" in the question and find the corresponding region in the image to make the right prediction. 
This indicates that our method can reduce the language priors, and make predictions by understanding the critical parts in the image-question pairs.

\textbf{Answer distribution.} 
To intuitively understand that our method overcomes the language priors effectively, we compare the answer distributions of some specific question types on VQA-CP v2 dataset in Figure \ref{answerdistribution}. We observe that the answer distribution of our method UpDn+DM is much closer to that of ground-truth answers from the test set than the baseline UpDn, while the answer distribution of UpDn mimics that of ground-truth answers from the training set to a greater extent. It indicates that UpDn+DM does not rely on the language priors in the training set that UpDn suffers from, but really understands the input to give answers on the test set. For instance, in the question type “is this”, “yes” makes the majority of answers given by UpDn+DM, as ground-truth answers from the test set; while UpDn mostly outputs the answer “no”, as ground-truth answers from the training set.

\textbf{Distance in the answer space.} As shown in Figure \ref{tsne}, we also visualize the distance between superficially similar instances in the answer space with t-SNE. We observe that with our method, superficially similar instances with the same question type and different ground-truth are separated more clearly in the answer space. 
It intuitively verifies that our approach successfully teaches the VQA model to distinguish between the superficially similar instances. 
To complement the t-SNE visualization, we also compute inter-class and intra-class Euclidean distances on answer classes with the same question type. The results for UpDn are 0.47 (intra-class distance in blue), 0.40 (intra-class distance in red), 0.52 (inter-class distance), and that for UpDn+DM are 0.32, 0.33, 0.50 respectively. It further demonstrates that our method's remarkable distinguishing ability is achieved by compressing the distance between intra-class data points.

\begin{table}[t]
\centering
\small
\setlength{\tabcolsep}{1mm}{\begin{tabular}{ccccc}
\toprule
Model & Overall & Yes/No & Num & Other \\
\midrule
\multicolumn{5}{l}{\textit{Model Ablation}}\\
\midrule
Ours w/o DM &  41.44 & 43.10 & 13.24 & 48.30\\
Ours w/o MF & 44.62 & 48.73 & 14.46 & 50.74\\
\midrule
\multicolumn{5}{l}{\textit{Data Ablation}}\\
\midrule
Ours w/o SC & 42.58 & 43.37 & 15.32 & 49.64 \\
Ours w/o RC & 60.47 & 89.19 & 39.39 & 51.21 \\
\midrule
Ours & 61.13 & 88.13 & 45.98 & 51.13 \\
\bottomrule
\end{tabular}}
\caption{Accuracies (\%) on VQA-CP v2 test set. "Ours w/o DM" denotes our model variant without the distinguishing module.
"Ours w/o MF" denotes our model variant without the modulating factor $p_{jm}$ in Equation \ref{focal}. "Ours w/o SC" and "Ours w/o RC" denote our model without using the synthetic counterparts and the real counterparts for training, respectively.}
\label{modulate}
\end{table}
\begin{table}[t]
\centering
\small 
\setlength{\tabcolsep}{1mm}{\begin{tabular}{lcccc}
\toprule
Model & Overall & Yes/No & Num & Other \\
\midrule
SAN\textsuperscript{\dag} & 28.60 & 37.64 & 11.51 & 28.54\\

+DM & 34.98 & 50.89 & 17.75 & 31.37\\
\midrule
UpDn\textsuperscript{\dag} & 41.44 & 43.10 & 13.24 & 48.30\\

+DM & 61.13 & 88.13 & 45.98 & 51.13\\
\bottomrule

\end{tabular}}
\caption{Accuracies (\%) on VQA-CP v2 test set based on the architecture of SAN and UpDn model.\textsuperscript{\dag} represents re-implementation results.}

\label{model agnostic}
\end{table}
\begin{table}[t]
\centering
\small
\resizebox{0.5\textwidth}{!}{
\begin{tabular}{cccccc}
\toprule
$N_1$ & $N_2$ & Overall & Yes/No & Num & Other \\
\midrule
1 & 1 & 61.17 & 88.15 & 46.08 & 51.16\\
1 & 2 & 61.20 & 88.55 & 45.82 & 51.08\\
2 & 1 & 60.91 & 88.18 & 45.28 & 50.91 \\
1 & 3 & 61.22 & 88.35 & 46.21 & 51.13\\
3 & 1 & 60.94 & 88.05 & 45.56 & 50.96\\
\midrule
\multicolumn{2}{c}{Average} & 61.09\textsuperscript{±0.15} & 88.26\textsuperscript{±0.20} & 45.79\textsuperscript{±0.38} & 51.05\textsuperscript{±0.11}\\
\bottomrule
\end{tabular}
}
\caption{Accuracies (\%) on VQA-CP v2 test set with different $N_1$ and $N_2$.}
\label{varyn1n2}
\end{table}
\begin{figure}[t]
    \centering
    \setlength{\belowcaptionskip}{-0.4cm}
    \includegraphics[width=0.5\textwidth]{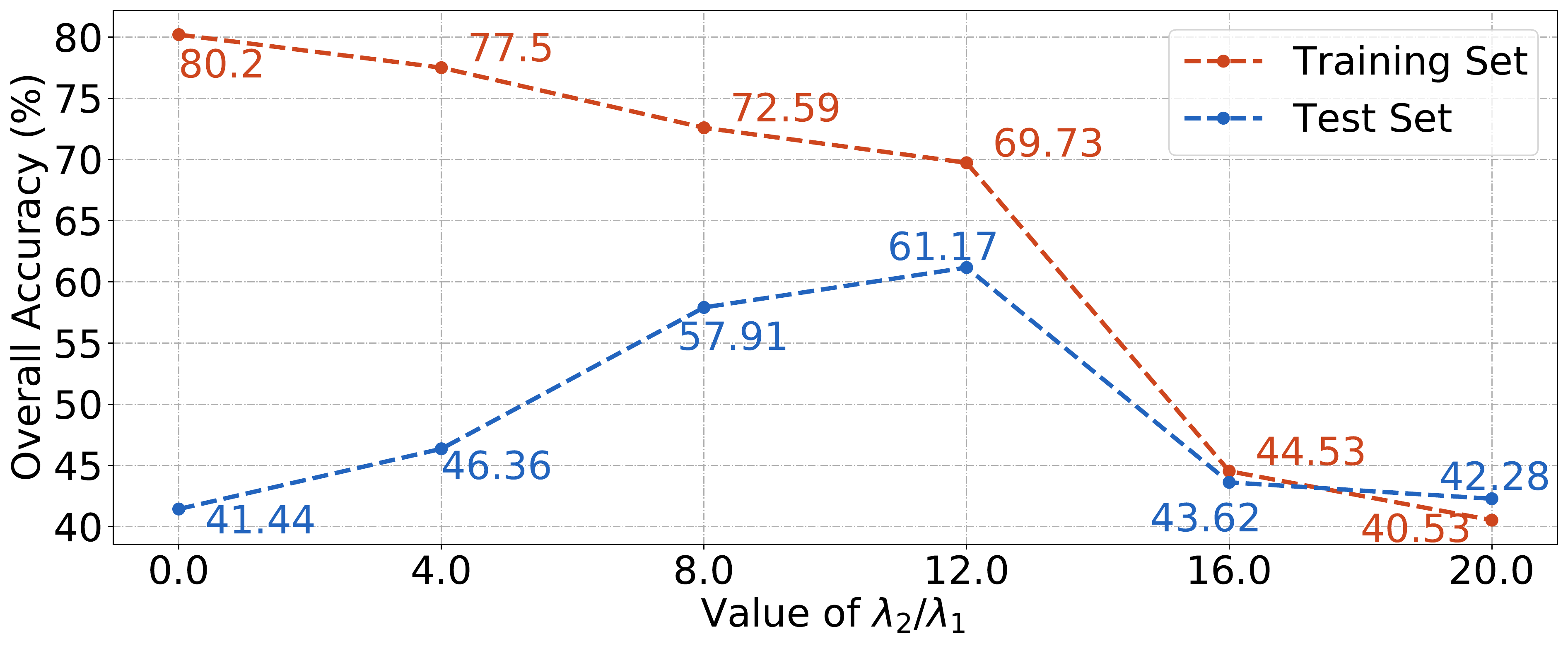}
    \caption{Overall accuracy (\%) on VQA-CP v2 training/test set when $\lambda_2$/$\lambda_1$ varies.}
    \label{lambda}
\end{figure}

\subsection{Ablation Study}
\textbf{Model ablation.}
In the first part of Table \ref{modulate}, we compare our method with its variant without the whole distinguishing module and the variant without the modulating factor $p_{jm}$ in Equation \ref{focal}. We observe that the model performance drops violently if we remove the whole distinguishing module or the modulating factor, which demonstrates that both of them are indispensable.

\textbf{Data ablation.}
As shown in the second part of Table \ref{modulate}, we observe that both the real counterparts and synthetic counterparts contribute to the improvement, and the synthetic kind contributes much more. This is reasonable since the real counterparts are far less than the synthetic counterparts in both quantity and diversity. The real counterparts only come from the VQA dataset itself with a limited size; while within the synthetic counterparts, a question can be combined with different images in different batches, which yields a large amount of diverse superficially similar instances. 

\subsection{Discussion}
\textbf{Model generalizability.} We further conduct experiments on different base models to validate the generalizability of our model. 
Besides the UpDn model, we also apply our method to the base model Stack Attention Networks (SAN) \cite{san}. As shown in Table \ref{model agnostic}, SAN+DM also achieves significant improvement in the accuracies of all categories compared to the original SAN, which shows our method can be generalized to other model architectures.

\textbf{Varying $N_1$ and $N_2$.} To investigate the influence of the sampled number of superficially similar counterparts in the resource-efficient implementation, we fix $N_1=1$ or $N_2=1$ and vary the other from 1 to 3. As shown in Table \ref{varyn1n2}, the results are stable with different $N_1$ and $N_2$. We set both parameters as 1 as default. \label{discuss_n1n2}

\textbf{Tuning the influence of $\bm{\mathcal{L}_{dis}}$.} We adjust the ratio $\lambda_2/\lambda_1$ to tune the influence of $\mathcal{L}_{dis}$. 
The results are shown in Figure \ref{lambda}. 
As the ratio of $\mathcal{L}_{dis}$ increases, the performance on the training set drops constantly, which demonstrates the model is less dependent on language priors to make predictions. 
Moreover, the performance on the test set first increases in the range of small $\lambda_2/\lambda_1$ values, and then decreases when $\lambda$ becomes too large. We deduce that if the ratio is too large, the influence of $\mathcal{L}_{dis}$ would overwhelmingly exceed that of $\mathcal{L}_{vqa}$ and make the VQA model underfit. Overall, we should set $\lambda_2/\lambda_1$ in a reasonable range to alleviate language priors as well as prevent underfit.

\section{Conclusion}
In this paper, we introduce the concept superficially similar instances and propose a novel training framework for overcoming language priors by explicitly encouraging the VQA model to distinguish between such instances. Extensive experiments show that our approach achieves the state-of-the-art results on VQA-CP v2, while maintaining competitive performance on VQA v2. Qualitative analysis also demonstrates that our method alleviates the language priors effectively and really understands the input.

\section*{Acknowledgements}
This research is supported by the Chinese Scientific and Technical Innovation Project 2030 (2018AAA0102100), the National Natural Science Foundation of China (No. 62272250, U1936206, U1903128).

\bibliography{ref_simplified}
\bibliographystyle{acl_natbib}

\end{document}